\documentclass[10pt, a4paper]{article}
\usepackage{lrec2022} 
\usepackage{multibib}
\newcites{languageresource}{Language Resources}
\usepackage{graphicx}
\usepackage{tabularx}
\usepackage{soul}
\usepackage{titlesec}
\titleformat{\section}{\normalfont\large\bfseries\center}{\thesection.}{1em}{}
\titleformat{\subsection}{\normalfont\SmallTitleFont\bfseries\raggedright}{\thesubsection.}{1em}{}
\titleformat{\subsubsection}{\normalfont\normalsize\bfseries\raggedright}{\thesubsubsection.}{1em}{}
\renewcommand\thesection{\arabic{section}}
\renewcommand\thesubsection{\thesection.\arabic{subsection}}
\renewcommand\thesubsubsection{\thesubsection.\arabic{subsubsection}}

\usepackage{epstopdf}
\usepackage[utf8]{inputenc}

\usepackage{hyperref}
\usepackage{xstring}
\usepackage{csquotes}

\usepackage{color}

\title{A Unified Approach to Entity-Centric Context Tracking in Social Conversations}

\name{Ulrich R\"uckert, Srinivas Sunkara, Abhinav Rastogi, Sushant Prakash, Pranav Khaitan} 

\address{Google Research \\
         1600 Amphiteatre Pkwy, Mountain View, CA 94043, USA \\
         \{rueckert, srinivasksun, abhirast, sush, pranavkhaitan\}@google.com\\}

\abstract{
In human-human conversations, Context Tracking deals with identifying important entities and keeping track of their properties and relationships. This is a challenging problem that encompasses several subtasks such as slot tagging, coreference resolution, resolving plural mentions and entity linking. We approach this problem as an end-to-end modeling task where the conversational context is represented by an entity repository containing the entity references mentioned so far, their properties and the relationships between them. The repository is updated turn-by-turn, thus making training and inference computationally efficient even for long conversations. This paper lays the groundwork for an investigation of this framework in two ways. First, we release Contrack, a large scale human-human conversation corpus for context tracking with people and location annotations. It contains over 7000 conversations with an average of 11.8 turns, 5.8 entities and 15.2 references per conversation. Second, we open-source a neural network architecture for context tracking. Finally we compare this network to state-of-the-art approaches for the subtasks it subsumes and report results on the involved tradeoffs. 
 \\ \newline  }

\begin{document}

\maketitleabstract

\section{Introduction}

Computers and mobile phones have changed how people communicate. A large amount of today's interpersonal communication happens in messaging apps on mobile devices or on chat and discussion services on the internet. Consequently, this has piqued the interest of the research community in developing assistive technologies for human-human conversations.
Representing the current status of a conversation in a succinct and semantically complete way is a central component of such technologies. 
At the core of this endeavor lies the task of tracking the entities mentioned in a conversation, their properties and the relationships that are being expressed about them. 
In this paper, we frame this task, which we call \emph{Context Tracking}, as an online machine learning problem, where the model is expected to track the current status of the conversation at any time. This formulation extends and complements existing research in three key areas.


First of all, in this framework the model ingests the messages of a conversation turn by turn and updates a growing repository of detected entity references in each turn. This allows for fast inference because the model only needs to ingest a single message and the repository instead of the entire conversation history. That is important particularly for long conversations, where ingesting the entire dialogue history is not feasible.

Second, the repository of entity references serves as a condensed storage of the semantic knowledge conveyed in the conversation. In this sense, context tracking serves the same purpose for open-domain conversations that dialog state tracking does for task-oriented conversations. While the current formulation only deals with tracking entity references, the same framework can be used to track relationships between entities or to build personal knowledge graphs. Once it can capture a large enough share of semantic information, the repository could be used to address higher-level tasks such as reasoning \cite{chen2020review}, question answering \cite{huang2019knowledgegraph} or summarization \cite{huang2020knowledgegraph}.

Third, since the framework aims at extracting a rich set of semantic information, it unifies existing NLP tasks such as entity recognition, slot tagging, coreference resolution and resolving plural mentions in one formulation. This means it is more parameter-efficient, because it shares parameters across multiple tasks. Recent research on large neural networks \cite{brown2020language,raffel2020exploring} has raised the prospect that larger networks with multiple endpoints may even improve predictive accuracy over single-task approaches.

In this paper, we take the first steps towards a systematic investigation of Context Tracking and lay the groundwork for further research. The most important part of this effort is the new Contrack dataset, which contains over 7000 social open-domain conversations with annotations for people and location entities with an average of 11.8 turns, 5.8 entities and 15.2 references per conversation. It is publicly available and can be downloaded at \url{https://github.com/google-research-datasets/contrack}.

In the second part of the paper we present a simple baseline model for Context Tracking. The model is a straightforward adaptation of a Transformer-based \cite{vaswani2017attention} neural network to the peculiarities of Context Tracking. Its conceptual simplicity makes it well suited to serve as a benchmark for further work. We release the code for this baseline implementation as Open Source. It is available at \url{https://github.com/google-research/google-research/tree/master/contrack}.

The third part of the paper reports the results of experiments we performed with the baseline model. These experiments give some insights on which parts of Context Tracking are solved easily and which parts require more research. We also compare the dataset and baseline model to other approaches for related tasks to get a sense of the tradeoffs involved with modeling multiple endpoints concurrently in turn-by-turn fashion.


\section{Related Work}
Context tracking or dialogue state tracking has been a well studied problem in task-oriented systems \cite{mrkvsic2017neural,rastogi2017scalable}, where the context is tracked in terms of the slots of the underlying task or API. However, a slot based representation is not suitable for human-human conversations because they tend to cover multiple domains making enumeration of slots impractical. Models like Meena \cite{adiwardana2020towards} and DialoGPT \cite{zhang2019dialogpt} model human-human conversation using a latent representation of the dialogue context. Such a representation is not suitable for applications where an explicit or interpretable representation of context is desired.

The availability of public datasets has played a central role in driving this area of research. Many datasets for coreference resolution have focused on either document based coreferences, like OntoNotes \citelanguageresource{ontonotes2013} (discussed by \newcite{pradhan2012conll}), ACE \cite{doddington2004ace} or conversations about specific tasks like MuDoCo \cite{martin2020mudoco} and CoQA \cite{reddy2019coqa}. Despite their scale and popularity, these datasets are not similar to human-human conversations as it is known that a large portion of human-human conversation centers on chit-chat, socialization and personal interests \cite{dunbar1997human}. Datasets like Persona Chat \cite{zhang2018personachat}, Daily Dialog \cite{li2017dailydialog} and others contain social conversations in different settings, but they are not suitable to be used for context tracking as they are not focused on references to entities in the conversations.

A variety of data driven techniques have been studied for semantic parsing and resolving ambiguities arising commonly in natural language \cite{yadav2019survey,lee2017end,sevgili2020neural}. With growing interest in Conversational AI and owing to the availability of public datasets \cite{chen2016character,gopalakrishnan2019topical,martin-etal-2020-camembert}, some of these techniques have been successfully applied to human-human conversations. \newcite{chen2016character} studied the problem of coreference resolution and linking mentions to a fixed set of entities, whereas \newcite{zhou2018they} investigated the problem of resolving plural mentions, which refer to multiple entities. Obtaining structured information from conversations has also been explored by \newcite{el2017nerex} for generating a graphical representation of entities mentioned in a conversation through pairwise co-occurence, and by \newcite{li2014personal} for constructing a user centric personal knowledge graph from conversations. 

These techniques separately address various important aspects of context tracking and representation of the entities mentioned in the conversation. Our framework combines the several important pieces of these works including coreference resolution, plural mentions, named-entity recognition, and attribute classification.

\begin{figure*}[ht]
\begin{center}
\includegraphics[width=\textwidth]{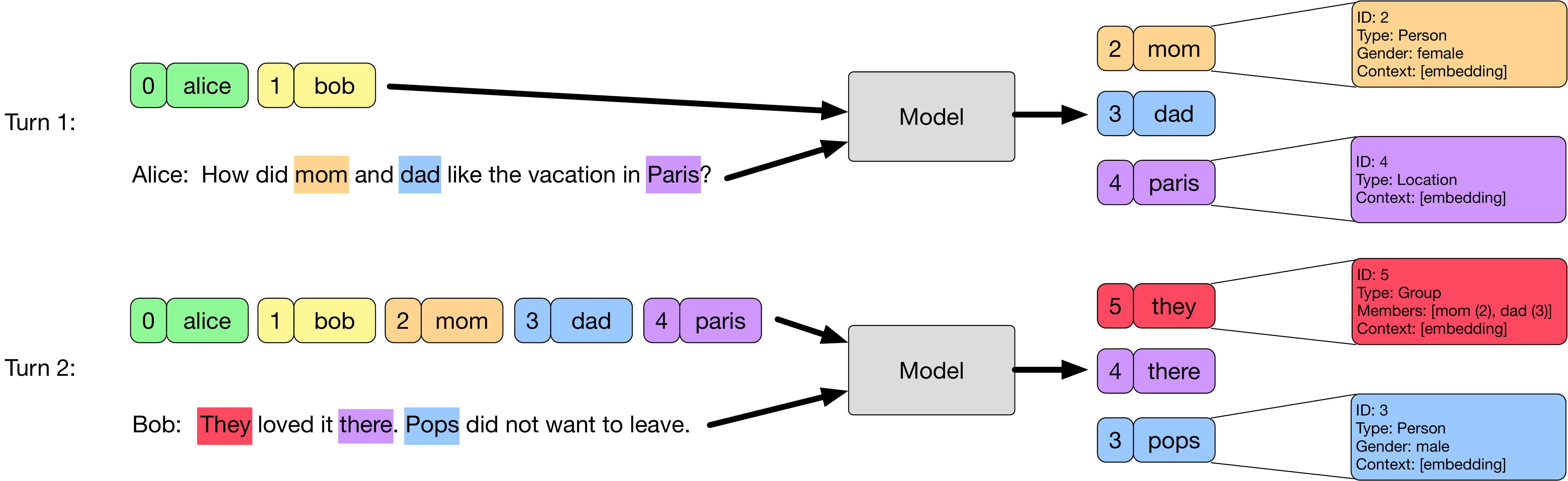}
\caption{Inspecting the contrack representation for two turns of an example conversation. The model is expected to identify tokens which refer to entities and outputs semantic data about them. The entity references output in turn one are passed as input to the model in the second turn.}
\label{example_figure}
\end{center}
\end{figure*} 

\section{The Context Tracking Modeling Task}
\label{modelingtask}

In the context tracking task the model processes conversations incrementally, one turn at a time. The model keeps track of the context within a repository of entity references. In order to disambiguate references in later turns, the repository also needs to store data about grammatical gender, group membership and other context data. In this section, we first describe the elements that are stored in the repository. Afterwards, we discuss the dataset we collected for this task.

\subsection{Context Representation}
The conversation context is represented by a repository containing all \emph{entity references} seen so far in the conversation. An entity reference is a span of tokens that refers to an entity. When adding an entity references to the repository we store the following information:

\begin{itemize}
    \item A unique \emph{entity ID}. This numerical ID uniquely identifies the entity or group of entities present in the entity repository. If two references refer to the same entity, they have the same entity ID. 
    
    \item A list of explicit \emph{entity properties}. Properties specify relevant  information about the entity such as the type of entity (e.g. people or location), grammatical gender, and whether the reference is a plural form. One could include other signals relevant for disambiguation or for downstream tasks that rely on this information.
    
    \item \emph{Group membership}. Some references are plural in nature, for example when referring to a group of people with \textit{they}. Such plural references (which we call \emph{groups} in the following) are added in the repository, and the group membership field enumerates the known entities which are members of a group.
    
    \item \emph{Implicit context data}. Additionally, the repository can optionally contain signals encoding information that is relevant to an entity but which cannot be easily captured by an explicit property. These signals could either be distributed (e.g. an embedding of surrounding tokens to the most recent reference) or discrete (e.g. a bag of salient words co-occurring with mentions of the entity). 
\end{itemize}

Figure \ref{example_figure} illustrates this representation on two turns of an example conversation between Alice and Bob about their parents' vacation. Before the conversation starts, the repository contains two entity references referencing the conversation participants Alice and Bob. When Alice sends the first message \textit{``How did mom and dad like the vacation in Paris?"}, the model identifies three new entities, one each for the person entities \textit{mom} and \textit{dad} and one for the location \textit{paris}. It serially assigns unique numerical IDs two, three and four to these new entities respectively, since the IDs zero and one are already taken. Note that the references also contain properties such as type and grammatical gender. In the next turn the repository contains five entity references. 

For the next message \textit{``They loved it there. Pops did not want to leave."}, the model is expected to identify three new entity references. The first one is \textit{they}, which is a group containing the two members \textit{mom} (ID two) and \textit{dad} (ID three). The references \textit{there} and \textit{pops} refer to existing entities and are thus labeled with the corresponding IDs. This annotation conveys that \textit{pops} and \textit{dad} refer to the same entity, as do \textit{there} and \textit{paris}.

\subsection{Dataset Collection and Details}
Real human to human casual conversations contain personal data and are hard to obtain. Hence, there is a need to collect a synthetic dataset which offers the same challenges as a real human-human conversation corpus. Our data collection procedure, outlined below, ensures that the conversations are natural, cover a variety of topics and mention different entities while minimizing annotation errors.

\subsubsection{Scenario Generation}
Our conversations are seeded by manually created scenarios which are short summaries of the content of the conversation to be generated. Here is a sample scenario:

\begin{displayquote}
\textit{Jeffrey's siblings, Nate and Marie are going to visit him and he asks Beverly for suggestions on where to take them. She asks him what they like to do and what they had done the last time they visited. She comes up with a few suggestions and they finally decide on a plan.}
\end{displayquote}
The scenarios are created by a separate group of expert crowd workers. They were asked to focus on variety and to leave certain details unspecified to facilitate creation of multiple conversations from each scenario. Scenarios enable us to control the broad topic of the conversation, to ensure that a variety of settings are captured and that relevant entities are mentioned. 

\subsubsection{Wizard-of-Oz Setup}
We developed a web application for collecting conversations for a given scenario by pairing two crowd workers together. For the last third of the dataset, we switched to using a single crowdworker for playing both roles in a conversation (similar to findings of \newcite{byrne2019}), as we found that it improved efficiency without a loss in quality. After an initial filtering of these conversations, the crowd workers annotated them by identifying spans which represent entity references. For each entity, the workers annotated the type of the entity and resolved all pronouns or noun phrases referring to this entity.

Over the course of the data collection for both tasks, we monitored the quality of data using heuristics like measuring the word overlap between the scenario and the chat, repeated turns for chat collection and comparing with known pronoun properties (e.g., the word “them” refers to a group etc.). We used these methods to update the list of good quality crowd workers who were used for subsequent data collection.

Finally, we had an additional round of verification on the annotated data to verify the annotations and to rate the chats on a scale of 1-5 for their naturalness and coherence. In addition chats with non-fluent English and grammatical mistakes not expected in a casual conversations were also given a low score. Chats with a score of 4 or 5 were retained and chats with a score of 3 were sent for rating by another crowd worker. Finally, we removed chats with scores less than 4 and we sent chats with annotation errors back for re-annotation. To increase the diversity of entities mentioned in our data and to prevent the model from overfitting on the names in the training set, we also release an augmented set of conversations in which we replace people names in the conversations with names randomly chosen from a very large collection.

\begin{figure}[t]
\begin{center}
\includegraphics[width=\columnwidth]{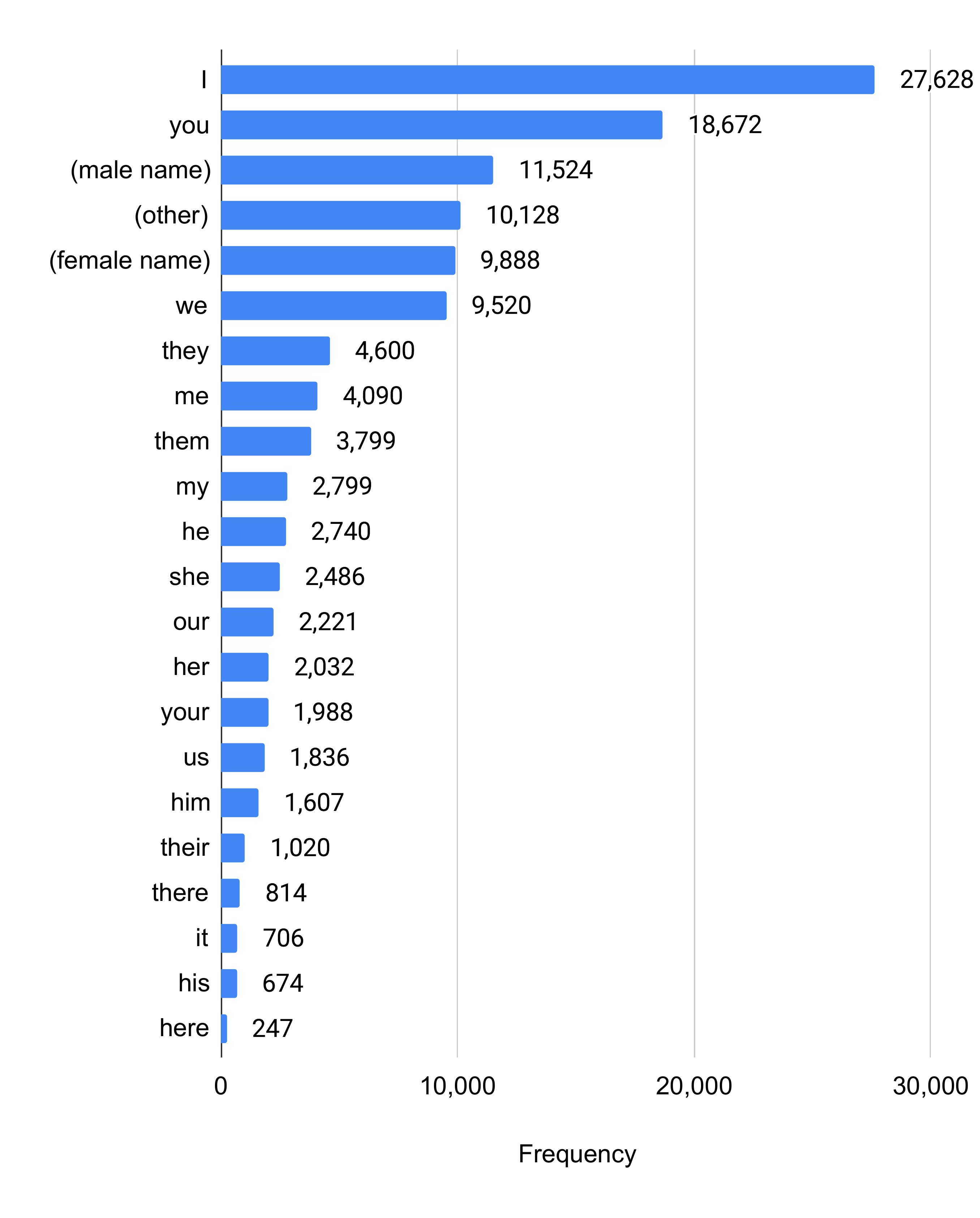}
\caption{The number of tokens spans which are most frequently annotated as entity references.}
\label{token_freq_figure}
\end{center}
\end{figure}

\begin{figure}[t]
\begin{center}
\includegraphics[width=\columnwidth]{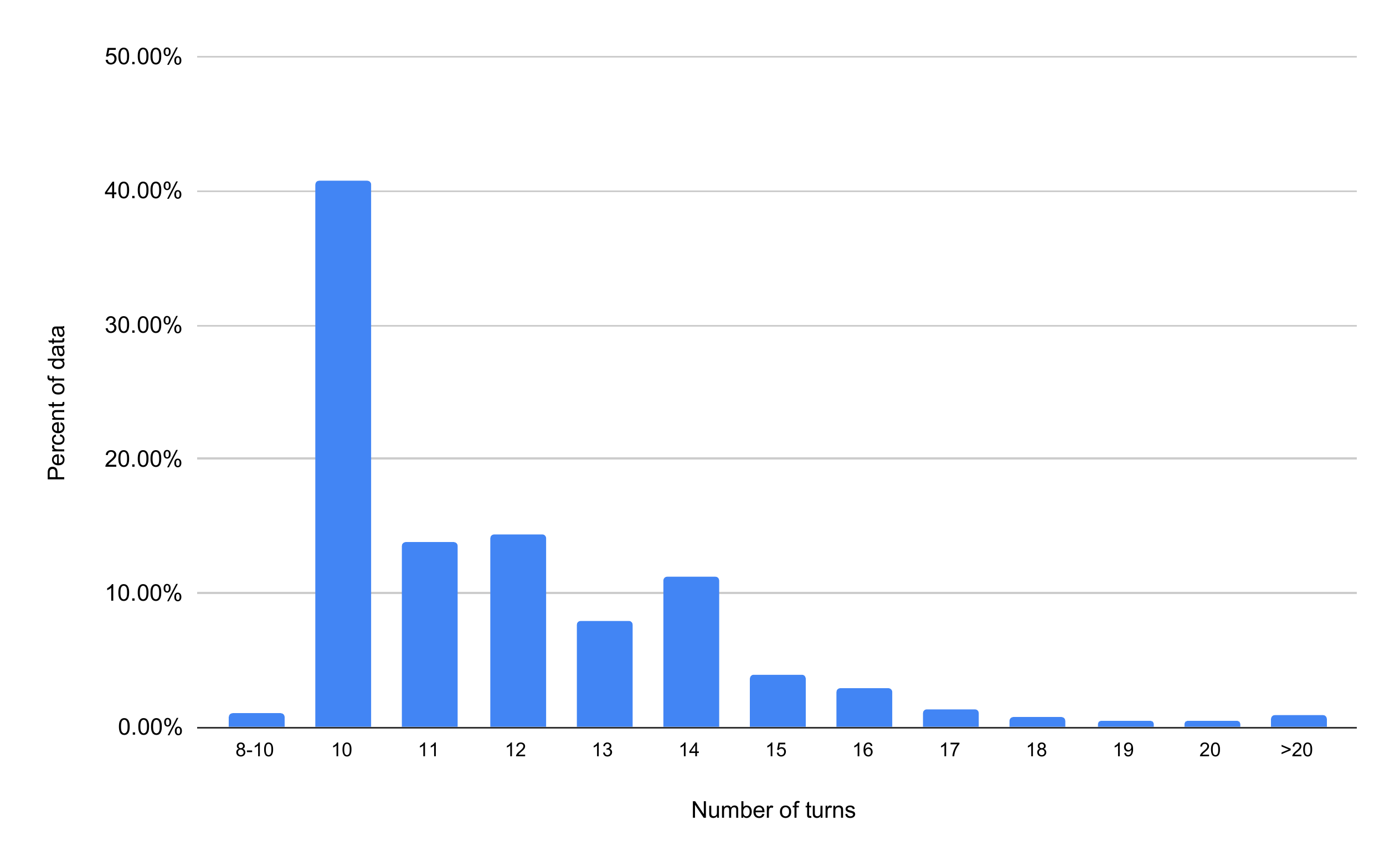}
\caption{Distribution of number of turns per conversation. The majority of conversations have 10-14 turns.}
\label{turns_distribution_figure}
\end{center}
\end{figure} 

\begin{figure}[t]
\begin{center}
\includegraphics[width=\columnwidth]{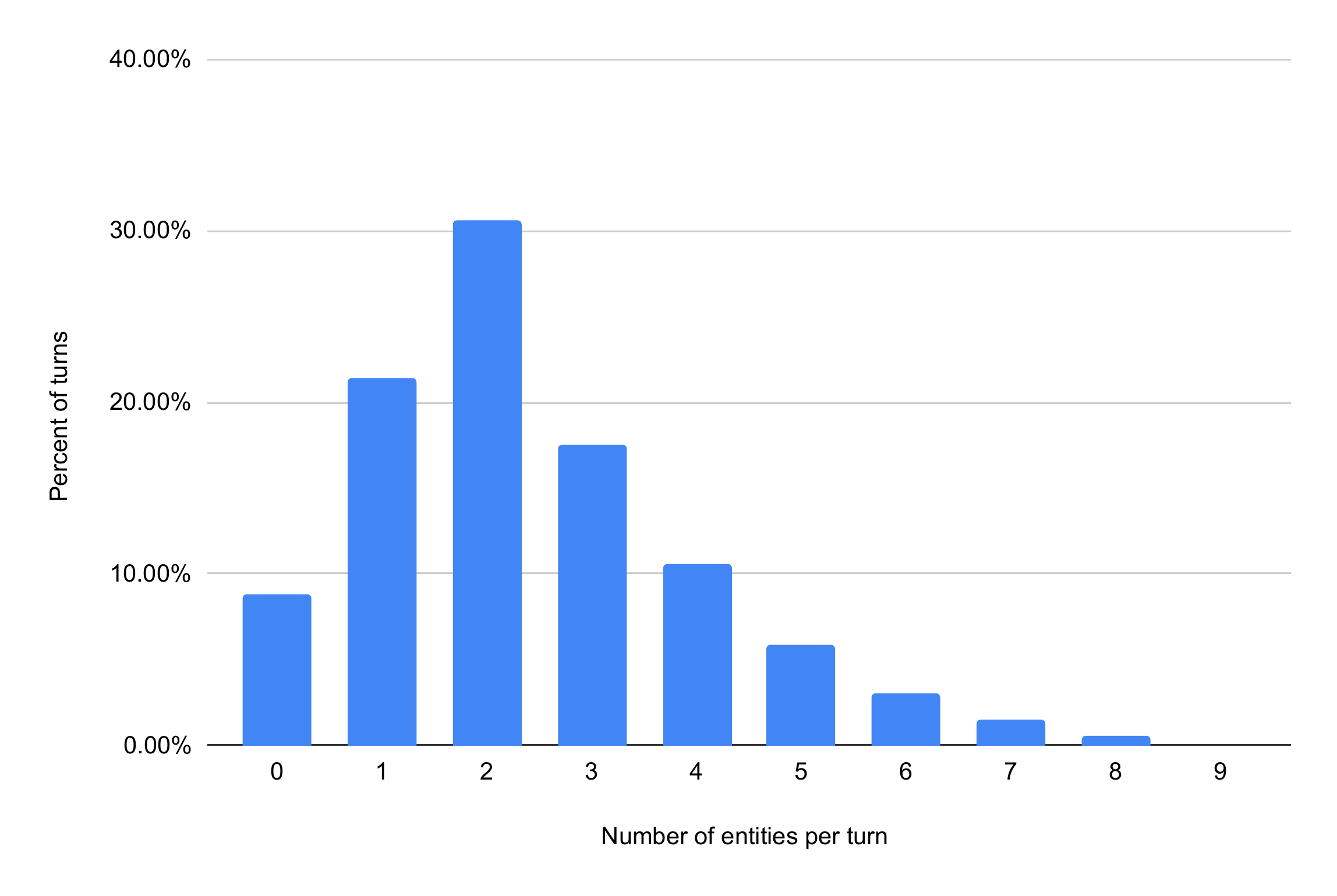}
\caption{Distribution of number of entities per turn. Around 8\% of the turns have no entities while 11\% of turns have 5 or more entities.}
\label{annotations_distribution_figure}
\end{center}
\end{figure} 

\subsubsection{Dataset Statistics}
The resulting Contrack Dataset is a human to human casual conversation dataset containing 7245 conversations with annotations for entities. Currently we have annotations for entities representing people and locations. The annotations contain signals for grammatical gender and group membership for people entities, and grammatical count and group membership for location entities. In total, we have 85,538 turns in the dataset with an average of 11.5 tokens per turn. Figure \ref{token_freq_figure} provides statistics of the most frequently annotated token spans in the dataset. Figures \ref{turns_distribution_figure} and \ref{annotations_distribution_figure} show the distribution of number of turns per conversations and the distribution of number of entities per turn respectively.

\section{The Contrack Baseline Model}
In this section, we present a simple model for context tracking to demonstrate the feasibility of the proposed approach and to act as a baseline for future studies on this dataset. The model takes an utterance and an entity repository as input, and outputs a list of entity references which will be added to the repository. One major challenge is the \textit{reference co-occurrence} problem, which occurs whenever a new entity is introduced and then referred to in the same turn by another token span (e.g. \textit{karen} and \textit{her} in ``\textit{have you heard karen totaled her car}"). It is a challenge because the entity ID for the reference \textit{her} needs to be the same as the entity ID for \textit{karen}, which has not been assigned yet. Our baseline model provides a simple solution to this problem, while jointly modeling all context tracking subtasks.

\subsection{Model Architecture}
\label{model_architecture}
The model solves the \textit{reference co-occurrence} problem by dividing the computation into two stages. The first stage identifies all references to new entities that have not been mentioned in the conversation so far. These newly mentioned entities are sequentially assigned new entity IDs, so that the rest of the model can refer to them in subsequent computation. For example, The second stage performs the rest of the tasks including assigning entity IDs to the other entity references, predicting properties of a reference (e.g. gender, plurality etc.) and group memberships for references referring to a group of entities (e.g. \textit{they}, \textit{us} etc.).

\begin{figure}[t]
\begin{center}
\includegraphics[width=\columnwidth]{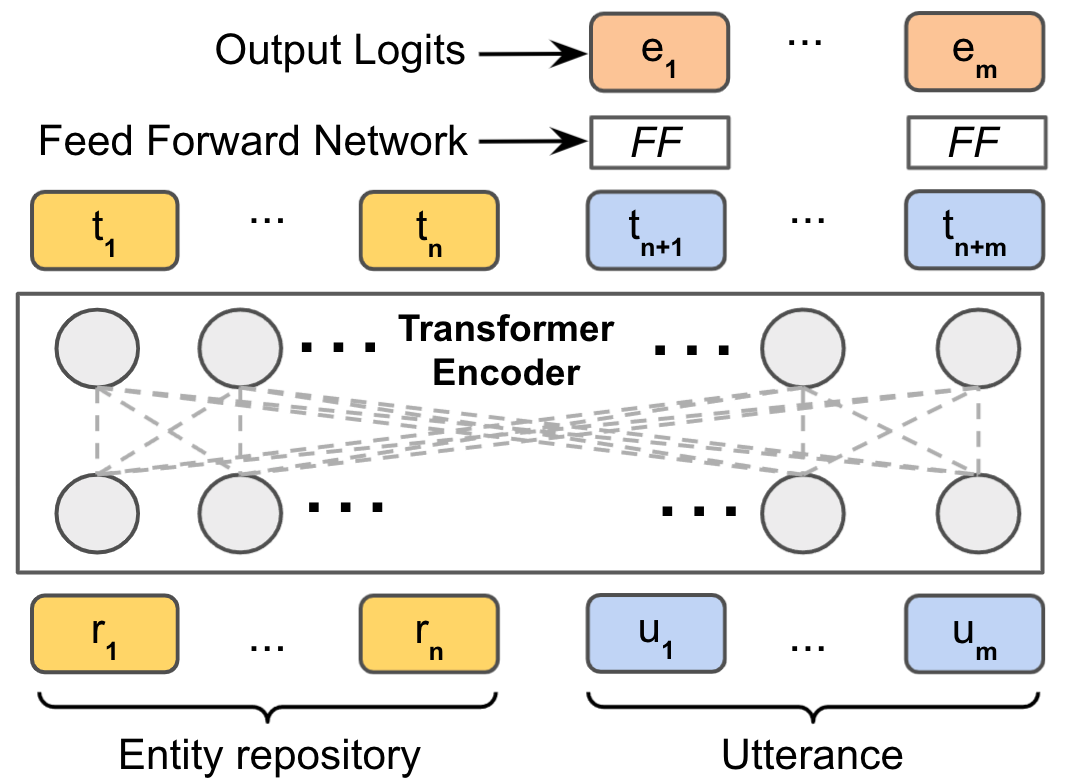}
\caption{Architecture for each of the two computation stages in the context tracking baseline model.}
\label{model_figure}
\end{center}
\end{figure}

Both stages have the same underlying architecture, which is shown in Figure \ref{model_figure}. The network takes two sequences as inputs. The first sequence $r_1, \ldots, r_n$ encodes the entity repository, where each element represents a single entity reference. The second sequence $u_1, \ldots, u_m$ corresponds to the input utterance, with each element refers to a token in the utterance. The vector representation of these sequences are obtained by concatenating the embedded representation of various features as described in the following section. The input sequences are fed to a 1-layer transformer encoder, which fuses the representations of the two sequences to generate the sequence $t_1, \ldots, t_{n+m}$. The outputs $t_{n+1}, \ldots t_{n+m}$ are then fed to a feed forward network to generate output logits $e_1, \ldots, e_{m}$ of appropriate dimension, from which all the required outputs are obtained as described in the Context Tracking section below. 

\subsection{Input Feature Representation}
\label{feature_representation}
The entity repository and utterance token sequences are represented by concatenating the embedded representations of the input features. Some features are only available for the entity reference repository, and hence are set to 0 for the token sequence. These features are: (i) \textit{ID} - It is a one-hot vector encoding the ID that has been assigned to an entity. (ii) \textit{Meta} - A Boolean vector indicating whether the reference introduces a new, previously unseen entity.  (iii) \textit{Properties} - Each categorical value of a property (e.g. gender=male) is assigned a Boolean value, and the Boolean representation for all the values for all the properties are concatenated together to represent the properties of an entity. (iv) \textit{Membership} - A Boolean vector which denotes which entities are members of the referenced group. The vector is zero for non-group mentions.  (v) \textit{Context} - This is a one-hot vector of size 2 indicating whether the underlying entity is the sender or recipient of the message and a binary vector indicating which of the preceding turns a reference was part of. 

A few other features are available for both the entity reference repository and the utterance token sequence. These are: (i) \textit{Type} - Boolean value indicating whether a sequence element is an entity reference or a token. (ii) \textit{Signals} - Boolean features which indicate if the underlying element belongs to a predefined lexicon. The reported model uses a single lexicon of first names. (iii) \textit{Word2Vec} - It contains the Word2Vec \cite{mikolov2013word2vec} embedding of the underlying token. For entity references, it contains the mean of the embeddings of the tokens which were marked as references to that entity in earlier turns. (iv) \textit{BERT} - It contains the BERT \cite{devlin2019bert} embeddings for each element. They are obtained by feeding the two sequences to a BERT encoder, and then taking the average of all wordpiece tokens corresponding to each element. The Word2Vec and BERT embeddings respectively provide a context free and context dependent representation of the element, both of which are important for context tracking.

\subsection{Context Tracking}
\label{output_calculation}
The context tracking model outputs logits which encode information about the entity references in the input utterance. These references are added to the entity repository to update it for the next turn. Each entity reference contains four types of labels: (i) \textit{Entity ID:} It clusters the references referring to the same entity together. (ii) \textit{Meta:} This is a Boolean label indicating whether a reference corresponds to the first mention of an entity. It can be used to identify the source entity among the ones which share the same entity ID. (iii) \textit{Properties:} This contains predictions for each of the labels such as gender, plurality, type etc. (iv) \textit{Group memberships:} For group references, it contains the entity IDs for all the member entities. All these features are converted to Boolean vectors similar to the procedure described in the preceding section to obtain ground truth labels without any loss of information. Next, we describe how the model predicts output labels in two stages and which loss function is used for training.

\paragraph{Stage 1 -} The first stage just identifies the spans corresponding to new entities in the input utterance. Each of the output logit vectors $e_i, 1 \leq i \leq m$, consists of two elements, which are converted to Booleans using a threshold of zero. The two Booleans respectively denote whether the underlying token is at the beginning or inside a span referring to a new entity. The tokens for which both these Booleans are zero are not part of a new entity span. Once these spans are found, the next available entity IDs are sequentially assigned to them, and their Meta attribute is also populated.

\paragraph{Stage 2 -} The second stage calculates the rest of the output labels. In addition to the inputs described in the Input Feature Representation section above, representations of the token level outputs obtained from the first stage (entity ID and Meta) are concatenated to the token vectors. A similar network as Stage 1 is then used to convert the input sequences to $2K + P$ output logits for each token. Here, $K$ is the maximum number of supported entities and $P$ is the total number of values taken by all the properties. These logits can be converted to Boolean vectors by thresholding at 0. Those logits specify the model outputs, with the first $K$ logits giving the entity IDs, followed by $P$ logits determining the value of each of the properties and the remaining $K$ logits giving group memberships. Please note that no span annotations are needed in this stage because the token level entity ID predictions identify the spans. This means one can simply merge successive tokens with the same predicted entity ID into a single reference.

\paragraph{Loss function -} Each logit $x$ and its corresponding binary label $y = \pm 1$ give rise to a single loss term using hinge loss, defined as $\max(0, 1 - xy)$. The overall loss is obtained by summing all of the loss terms arising from each of the logits for each token, while masking out the summands arising from labels, which do not need to be considered at that particular position. For example, the logits for group membership predictions contribute to the loss only when the underlying reference is plural in the ground truth. Since the two stages are trained jointly, the overall loss $L$ is defined as $\alpha L_1 + L_2$, where $L_1$ and $L_2$ are the losses of the two stages and $\alpha$ is a tunable parameter. Empirically, $\alpha = 6$ gave us the best results, which indicates that misidentifying new entities should have a higher penalty.

\subsection{Implementation Details}
All the experiments use the same hyperparameters, with the transformer encoder made up of a single self-attention layer with 9 heads followed by a feed-forward network with a hidden-dimension of 800. The model is configured to track up to twenty entities ($K = 20$) and the input sequence pair is restricted to be no longer than 100 elements altogether. We do not use dropout during training and utilize Adam optimizer \cite{kingma2014method} with a learning rate of 0.0001 and batch size 20. 

The context tracking model is recurrent because the entity reference repository output in the current turn is used as an input for the next turn. We resolve this dependency during training by applying a teacher-forcing approach \cite{goyal2016professor} for simplicity, where the model uses ground truth labels as input instead of its output in the last step. Since the distribution of the entity IDs is skewed towards the smaller numbers in the dataset, we randomize the entity IDs in the training examples to make the model more robust. This is done by adding a single random number to all entity IDs in a conversation and taking modulo $K$ to keep the IDs within the supported range.

\begin{table}[t!]
\begin{center}
\begin{tabular}{|l |  c |c |c |} 
\hline
Section & P & R &F1 \\
\hline \hline
New Entities &  76.1  &  78.8  &  77.4    \\
Existing ID &  66  &  63.7  &  64.8    \\
Properties &  98.9  &  98.8  &  98.8    \\
Membership &  79.6  &  81.7  &  80.6    \\
\hline
\end{tabular}
\end{center}
\caption{Results for the four endpoints of the Contrack baseline model.}
\label{results_table}
\end{table}

\begin{figure*}[t]
\begin{center}
\includegraphics[width=\textwidth]{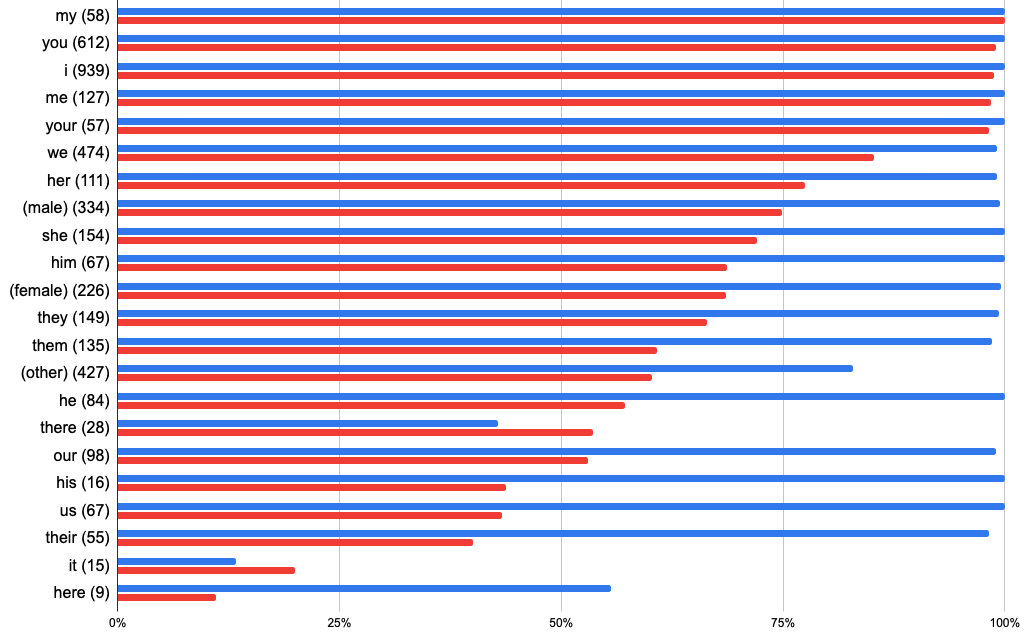}
\caption{Accuracy numbers for frequently occurring entity references. The blue bars give the percentage of entity references that were detected correctly, the red bars give the percentage with correctly predicted entity IDs.}
\label{accuracy_per_token_figure}
\end{center}
\end{figure*}

\section{Experiments}
\label{experiments_section}
First, we conduct several experiments to study the performance of our baseline model on the Contrack dataset. Next, we compare the baseline model to high performing task-specific models for a few conversational understanding tasks subsumed by context tracking.

\begin{table}[t!]
\begin{center}
\begin{tabular}{|c |  c |c |c |} 
\hline
Nr of & With Teacher & W/o Teacher & Difference \\
Turns & Forcing & Forcing & \\
\hline \hline
0 & 88.9 & 88.9 & 0.0 \\
1 & 84.6 & 78.3 & 6.3 \\
2 & 84.4 & 67.8 & 16.6\\
3 & 79.6 & 61.3 & 18.2\\
4 & 77.8 & 61.4 & 16.4\\
5 & 76.2 & 52.1 & 24.1 \\
6 & 75.3 & 54.6 & 20.7 \\
7 & 80.7 & 59.3 & 21.4 \\
8 & 79.8 & 58.7 & 21.2 \\
9 & 82.5 & 62.1 & 20.3 \\
\hline
\end{tabular}
\end{center}
\caption{Results on how error propagation affects prediction accuracy by turn. The columns give the percentage of correctly predicted entity IDs with and without Teacher Forcing.}
\label{errorpropagation_table}
\end{table}

\begin{table}[t!]
\begin{center}
 \begin{tabular}{l | c | c | c} 
 \multicolumn{4}{c}{\textbf{Attribute Classification}} \\
 \hline
 & Precision & Recall & F1 \\
 \hline
 Coach & 98.2 & 97.8 & 98.0 \\
 Coach-B & 98.2 & 98.0 & 98.1 \\
 Contrack & \textbf{99.0} & \textbf{99.0} & \textbf{99.0} \\
\hline
\multicolumn{4}{c}{} \\
  \multicolumn{4}{c}{\textbf{Singular Coreference Resolution}} \\
 \hline
 & Precision & Recall & F1 \\
 \hline 
 E2E Coref & \textbf{97.6} & \textbf{87.3} & \textbf{91.5} \\
 Contrack & 85.5 & 70.4 & 76.9 \\
\hline
\multicolumn{4}{c}{} \\
 \multicolumn{4}{c}{\textbf{Plural Coreference Resolution}} \\
 \hline
 & Precision & Recall & F1 \\
 \hline
 CZC & \textbf{81.2$\pm$1.0} & 73.3$\pm$0.4 & 75.9$\pm$0.5 \\
 TE & 80.4$\pm$0.8 & \textbf{76.5$\pm$1.2} & \textbf{78.0$\pm$0.6} \\
 Contrack & 73.2$\pm$0.6 & 68.5$\pm$0.7 & 69.7$\pm$0.8\\
\hline
\multicolumn{4}{c}{} \\
 \multicolumn{4}{c}{\textbf{Entity Linking}} \\
 \hline
 & Precision & Recall & F1 \\
 \hline
 CZC & \textbf{71.8$\pm$0.4} & 61.4$\pm$0.4 & 66.2$\pm$0.4 \\
 TE & 71.1$\pm$0.4 & \textbf{64.2$\pm$1.3} & \textbf{67.4$\pm$0.8} \\
 Contrack & 61.2$\pm$1.0 & 55.5$\pm$1.6 & 58.2$\pm$1.2 \\
 \hline
 \end{tabular}
\end{center}
\caption{Experiments: Comparison with various task specific SoTA models. Contrack is the baseline described in the Contrack Baseline Model section, Coach model is from \protect\cite{liu2020coach}, Coach-B is the Coach model trained with BERT embeddings, CZC is from \protect\cite{chen2017robust} and TE is from \protect\cite{zhou2018they}. More details can be found in the Experiments section.}
\label{other_tasks}
\end{table}

\subsection{Evaluation on Context Tracking}
The evaluation dataset is constructed by randomly setting aside 15\% of the data. This is done while ensuring that all conversations stemming from the same scenario are either assigned to the training or the evaluation set, but not split between the two. Unlike training, which uses teacher forcing, the model uses the predicted entity reference repository in a given turn as an input to the next turn. This makes the model more computationally efficient by allowing it to process each turn only once, and doesn't put any theoretical bound on the number of turns in a conversation. However, it also makes the model prone to error propagation because of a recurrent dependency.

We evaluate the performance of the two stages of model on four endpoints - (i) Identifying \textit{new entities}, (ii) Associating entities to \textit{existing IDs}, (iii) Predicting {properties} for all entities, and (iv) Calculating \textit{membership} of group entities. The first endpoint is implemented as part of the first stage, the others are computed by the second stage. We report the precision, recall and F1 score for each of these endpoints. 

Table \ref{results_table} lists the value of evaluation metrics.
They indicate that the model is generally accurate on properties and to a lesser degree on group membership, while detecting new entities is easier than resolving existing ones. That is true in particular for location entities where the model is generally not able to detect entities as reliably as with people. Figure \ref{accuracy_per_token_figure} shows the accuracy results for the most frequent occurring tokens. Not surprisingly, references to the first and second persons are the easiest to resolve while references to locations and third persons are more challenging. All in all, the baseline detected $96.8\%$ and resolved the IDs of $64.7\%$ of references correctly.

\subsection{Effects of Error Propagation}
Before we compare the baseline model's performance on other tasks, let us try to quantify the effect of error propagation within Context Tracking. Processing the conversation turn-by-turn can leads to cases where misprediction in one turn can result in errors in all subsequent turns. This makes Context Tracking more difficult than tasks where the model can use the whole conversation. To evaluate the magnitude of this effect, we compare the percentage of correctly predicted entity IDs of the baseline Contrack model with propagating entity repositories turn by turn (i.e. without teacher forcing) with a version which uses the ground-truth input repository as an input in each turn (i.e. with teacher forcing). The results are listed in Table \ref{errorpropagation_table}. We can see that the performance of the model drops as the index of the turn in a conversation increases. The difference is largest at five turns with a 24\% accuracy difference and remains flat around 20\% as the turn number increases.

\subsection{Evaluation on Other Tasks}
In this section, we demonstrate the applicability of the baseline model to conversational understanding tasks subsumed by context tracking. In all experiments we use a fully trained baseline model, which makes predictions for all endpoints. Its performance is compared to the target model trained on the single task under consideration. 
The baseline differs in that it is more parameter-efficient (because it models multiple endpoints concurrently) and easier to scale (because it makes predictions turn by turn). However, unlike the competing systems, the baseline's errors in one turn will propagate to later turns as described in the section on error propagation above. This means the results in this section can be considered to be a measure of the accuracy costs incurred by switching to turn-by-turn prediction and by using a joint model.

\paragraph{Slot Tagging.}
The slot tagging task \cite{goo2018slot,zhang2019dialogpt} deals with identifying spans from an utterance and associating a type or slot to these spans. The baseline model subsumes slot tagging by identifying the entity reference tokens as spans and assigning a type to it by predicting a property. We benchmark the baseline model against Coach \cite{liu2020coach}, which is one of the recent, best performing models for slot tagging. We train this model on the Contrack dataset, where it is only asked to identify the spans corresponding to entity references, and to assign their properties (grammatical gender for person entities, count for location entities). All other annotations are ignored. Table \ref{other_tasks} shows the results using the default fastText \cite{bojanowski2017fasttext} embeddings used by the baseline and BERT embeddings used by our model. Even though both models perform very well on this task, our simple baseline outperforms the Coach model in both settings.

\paragraph{Singular Coreference Resolution.}
We compare with the state of the art E2ECoref system \cite{lee2018higherorder} for coreference resolution, trained using default parameters on the Contrack dataset with all the plural entity references removed. Table \ref{other_tasks} gives the results. The E2ECoref system outperforms the baseline substantially. This can be partly explained by the fact that E2ECoref processes the entire conversation together, and hence has access to future turns while making a decision. As mentioned above the Contrack baseline makes its decisions based only on the current utterance and outputs from the previous turn and is thus prone to error propagation. However, future work may show that one can transfer some of the techniques which make E2ECoref successful to context tracking. 

\paragraph{Plural Coreference Resolution.}
We compare with \cite{zhou2018they}, which reports results on resolving plural mentions on conversational data. For the comparison, we train and evaluate the Contrack baseline model on the Character mining dataset \footnote{The publicly available code for the system described in \cite{zhou2018they} is missing some parts, so we were not able to train it on Contrack data.}, which contains annotated transcripts of four seasons of the TV show \textit{Friends}. We follow the same setup as \cite{zhou2018they}, computing mean and standard deviation of the BLANC score \cite{recasens2011blanc} over five runs. Results in Table \ref{other_tasks} show that the Contrack model's F1 score is lower by about eight points, which indicates that the baseline is more competitive on plural than on singular coreference resolution.

\paragraph{Entity Linking.}
\newcite{zhou2018they} also discusses entity linking, which differs from coreference resolution in that the model needs to assign each mention to one of the given known entities (that is, TV show characters in the Character Mining dataset). This is conceptually close to Context Tracking, but differs in scope (only people mentions), setup (entities are known in advance) and difficulty (model uses whole conversation). We compare the baseline model with the entity linking implementation in \cite{zhou2018they} as described in the previous section and report F1 scores in table \ref{other_tasks}. The F1 scores are worse by 9 points, similar to plural coreference resolution. On this dataset most of the loss is caused by plural references whose F1 score is fifteen points worse than the one reported by \newcite{zhou2018they}.

\section{Conclusion}
In this paper we introduce a new machine learning framework which tracks the entities in social open-domain conversations turn by turn, thereby building a repository of rich semantic data on the entities, their properties, and entity relationships for plural mentions. To enable research on the merits of this framework, we release the Contrack dataset and describe the implementation of a baseline model. While the experimental results are encouraging, it is clear that more work is necessary to turn context tracking into a mature practically useful service. 

There are a few directions for future research: For one, we plan to extend the dataset with more conversations and more annotations for multiple domains. This would allow us to track a more complete set of entities more reliably. On the other hand, more research is necessary to investigate which neural network architectures are best suited for accurate context tracking.

\label{sect:pdf}
\section{Bibliographical References}\label{reference}
\bibliographystyle{lrec2022-bib}
\bibliography{main}

\section{Language Resource References}
\label{lr:ref}
\bibliographystylelanguageresource{lrec2022-bib}
\bibliographylanguageresource{languageresource}

\end{document}